\documentclass{article}
\usepackage{ijcai21}

\usepackage{times}
\usepackage{soul}
\usepackage{url}
\usepackage[hidelinks]{hyperref}
\usepackage[utf8]{inputenc}
\usepackage[small]{caption}
\usepackage{graphicx}
\usepackage{amsmath}
\usepackage{amsthm}
\usepackage{booktabs}
\usepackage{algorithm}
\usepackage{algorithmic}
\title{Communication Optimization in Large Scale Federated Learning using Autoencoder Compressed Weight Updates}

\author{
Srikanth Chandar$^{1,2}$
\and
Pravin Chandran$^1$\and
Raghavendra Bhat$^1$\And
Avinash Chakravarthi$^1$
\affiliations
$^1$Intel Corporation, Bangalore\\
$^2$PES University, Bangalore
\emails

srikanth.chandar@gmail.com,
\{pravin.chandran, raghavendra.bhat, avinash.chakravarthi\}@intel.com,
}

\begin{document}

\maketitle

\begin{abstract}
Federated Learning (FL) solves many of this decade's concerns regarding data privacy and computation challenges. FL ensures no data leaves its source as the model is trained at where the data resides. However, FL comes with its own set of challenges. The communication of model weight updates in this distributed environment comes with significant network bandwidth costs. In this context, we propose a mechanism of compressing the weight updates using Autoencoders (AE), which learn the data features of the weight updates and subsequently perform compression. The encoder is set up on each of the nodes where the training is performed while the decoder is set up on the node where the weights are aggregated. This setup achieves compression through the encoder and recreates the weights at the end of every communication round using the decoder. This paper shows that the dynamic and orthogonal AE based weight compression technique could serve as an advantageous alternative (or an add-on) in a large scale FL, as it not only achieves compression ratios ranging from 500x to 1720x and beyond, but can also be modified based on the accuracy requirements, computational capacity, and other requirements of the given FL setup.
\end{abstract}

\section{Introduction}

\label{submission}
Federated Learning has been proposed as a new learning paradigm to overcome the privacy regulations and communication overheads associated with central training \cite{mcmahan2017communication},\cite{li2020federated}. In this paper, for sake of clarity, we assume an FL scheme where a central server (called Aggregator) shares a global model with participating edge devices (called Collaborator), and the model is trained on the local datasets available at the edge device. The local dataset is never shared with the central server, instead, local updates to the global model are shared with the central server. The central server combines the local updates from the participating clients using an Optimization or Aggregation Algorithm (FedAvg \cite{mcmahan2017communication}, FedProx \cite{li2018federated}, FedMa \cite{wang2020federated}, FedMAX\cite{chen2020fedmax}) and creates a new version of the global model. This process is repeated for the required number of communication rounds until the desired convergence criteria are achieved. The chosen scheme does not limit the technique from being deployed in any alternative FL scheme.

In the proposed FL setup, an AE compresses the weight updates from each collaborator at every communication round to reduce the communication bandwidth and recreates the weights after communication during the aggregation stage at the Aggregator. It is well understood that the weight updates are non-isolated events and there is some relation and interlinking between the parameters of the weight updates. A neural network such as an AE can find such patterns hidden in the data and reduce the representation to a lower-dimensional feature size (compression). 

\begin{figure}[ht]
\vskip 0.2in
\begin{center}
\centerline{\includegraphics[width=\columnwidth]{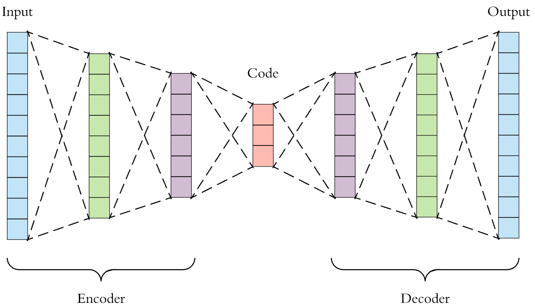}}
\caption{Autoencoder (Encoder and Decoder) Model}
\label{Fig AE}
\end{center}
\vskip -0.2in
\end{figure}

This paper investigates and shows that an AE can exploit this fact of dependence between parameters in a weight update, and thus learn the encoding of the weight update, and replicate it in a "learnable" manner. An AE model has a funnel structure where the hidden layers have neurons lesser in number compared to the input and output layers. The idea is that the network is trained to replicate the inputs, at the output level, by having certain intermediate lower-dimensional representations in the hidden layers. This "encoded" weight update, being smaller in size, is then communicated to the server- thus reducing the communication overheads. 

\section{Past Work}
Communication efficient FL has gained considerable attention over the past few years, and several approaches towards compressing the collaborator-to-server and server-to-collaborator model updates have been proposed. Traditional methods like sub-sampling, and quantization as implemented in \cite{reisizadeh2020fedpaq} compress the model update by the process of mapping weight parameter values to a smaller set of discrete finite values. Communication overheads can further be reduced by a certain more sophisticated quantization approach by combining the universal vector quantization, following a similar fundamental principle \cite{shlezinger2020federated}. Sub-sampling and Quantization are simplified methods that do not capture the inherent nonlinearities in the model while performing a compression but rather performs an approximation to reduce the size of the weights being communicated. Pruning reduces the model size by modifying architecture. Not all weights in a model are considered important, and an importance metric is defined for each of the weights. This metric is decided based on how each of these parameters causes the loss function to change when faced with a gradient. Unimportant weights are assumed a weight equal to zero, implying they can be removed from the network, finally resulting in a smaller network.

Federated Dropout \cite{caldas2018expanding} proposes a compression scheme derived from dropout regularization, where a fixed number of activations are dropped at each connected layer. This ensures local updates have reduced architecture and thus reduces communication bandwidth. DGC (sparsification)\cite{lin2017deep} only communicates the weights above the set threshold, and the others are accumulated locally on the device till it reaches the threshold value, and is only then communicated. The reasoning being 99\% of the update are redundant since they get altered during future communication rounds. CMFL \cite{luping2019cmfl} introduces an orthogonal compression method making use of the global model update tendency. The weights of the local updates that do not align with this tendency are not communicated, as they are deemed to have low importance since they would get corrected as the communication rounds proceeds. STC \cite{sattler2019robust} is similar to DGC, and includes compression for server-to-client communication, through the modification of the update rule to include client-side and server-side updates. FetchSGD \cite{rothchild2020fetchsgd} uses sketching and streaming to compress weight updates by summarizing them through a linear sketching algorithm (Count Sketch), using the accumulator and momentum variables from update equations stored on the server, instead of locally. 
Fedboost \cite{hamer2020fedboost} treats the FL weight update compression problem as an ensemble-based training optimization problem, by considering each of the collaborators as a member of the ensemble and aggregating the inferences instead of weights. The drawback is that it is applicable for density estimations only, and does not solve the weight update compression in itself. Similarly, Fedzip \cite{malekijoo2021fedzip} proposes a quantization approach using K-Means-Clustering, aided with sparsification and encoding, to compress the weight updates. Quantization through clustering provides a better reflection of tensor distribution, compared to other methods where a fixed set is selected. 

In Knowledge Distillation \cite{hinton2015distilling} the collaborator model is translated to a smaller model that replicates the original model behavior to a satisfactory level of accuracy, and the communication of weight updates during each communication round is thus of a smaller size \cite{seo2020federated}. The drawback of this approach is that the smaller model may not provide the entire information encapsulated in the collaborator model. Even with stronger teacher models \cite{lan2018knowledge}, there is a hesitation in commercially adopting this approach, as there could be sub-optimal distillation if the student model is too small and is over-fitting. There is also a cap on the amount of compression that may be achieved, as the process is dependent on recreating the collaborator model. 
Deriving from the idea of compressing the collaborator model weights, this paper proposes an architecture that builds a model that recreates the weight updates at each communication round via compression through an autoencoder, instead of recreating the collaborator model itself. While the AE model may be bigger than than the student model, the compression achieved through the AE would be significantly higher as it learns the patterns between the parameters of the weight updates, rather than between the input and output of the collaborator model (which ideally necessitates a model to be as big as the collaborator model, and hinders a wide range of compression). AEs have been used in the image compression domain \cite{zebang2019densely}. In the context of compressing another neural network using an AE, there is limited investigation. The closest research along these lines is the ALF \cite{frickenstein2020alf} approach, proposed for the embedded system track. Here, an AE is used to compress the weights at the end of every convolutional layer, to save memory usage while storing intermediate weights on the hardware. Whereas in the AE-based compression proposed in this paper, the compression occurs for all layer weights (convolutional or not) and is for the entire model. This provides the advantage of allowing the AE to learn the dependencies of the parameters across layers, and perform compression accordingly. This proposed method is suggested as a more dynamic approach serving as an advantageous add-on for communication efficient federated learning, by also considering the various trade-offs. 

\section{Architecture}

\begin{figure}[ht]
\vskip 0.2in
\begin{center}
\centerline{\includegraphics[width=\columnwidth]{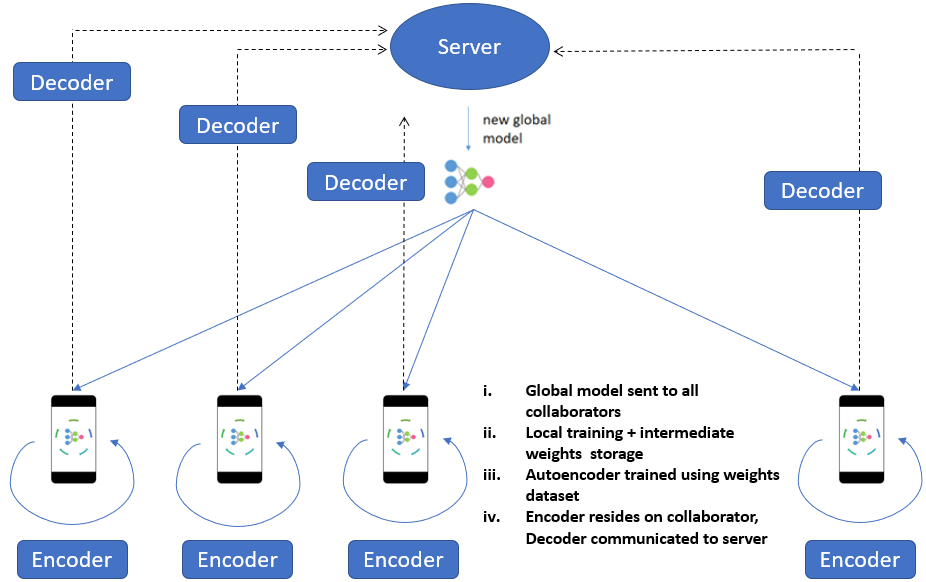}}
\caption{Prepass Round}
\end{center}
\vskip -0.2in
\end{figure}
The compression scheme proposed in this paper works in a manner where an AE neural network is trained to recreate the weight updates via an intermediate compressed state. To train the AE to compress a model, the weights data at the end of every batch/epoch of the training model (to be compressed) is required, as the AE learns the encodings from such a dataset. This implies that in the FL setup, a pre-pass round is required to generate the dataset containing weights associated with periodic learning. This is achieved as such - The server initiates the learning process on all collaborators by communicating a global model. Each collaborator trains this particular model using the local data available on the device. At this stage, the learning process happens without aggregation of weights or Federation. While the collaborators' models are being trained locally on-device data, the intermediate weights (at the end of every batch/epoch) are stored to form the weights dataset. The weights dataset generated at every collaborator is used to train an AE at the collaborator, in such a way that the AE can reproduce the weights, and also compress them through its intermediate smaller dimensional layer. The decoder part of the AE is communicated to the server, which also concludes the pre-pass round. Beyond this stage, the FL is initiated by the server using a global model. This time each collaborator starts training on the local data, but with aggregation and federation at every communication round. The weight updates from the collaborator to the server are compressed using the encoder on the collaborator and reconstructed using the decoder on the server. The compression ratio achieved per communication round, per collaborator is around 1500x, and this factor can be modified based on the AE structure. 
The encoding network can be represented by the standard neural network function passed through an activation function, where z is the latent dimension.
\begin{equation}
\small
z= \sigma(Wx + b)
\end{equation}
Similarly, the decoding network:
\begin{equation}
\small
x'= \sigma'(W'z + b')
\end{equation}
The loss function in terms of these network functions used to train the neural network through the standard backpropagation procedure is:
\begin{equation}
\small
\mathcal{L}(x,x')=\| x-x' \|^2 = \| x-\sigma'(W'\sigma(Wx + b) + b') \|^2
\end{equation}

\begin{figure}[ht]
\vskip 0.2in
\begin{center}
\centerline{\includegraphics[width=\columnwidth]{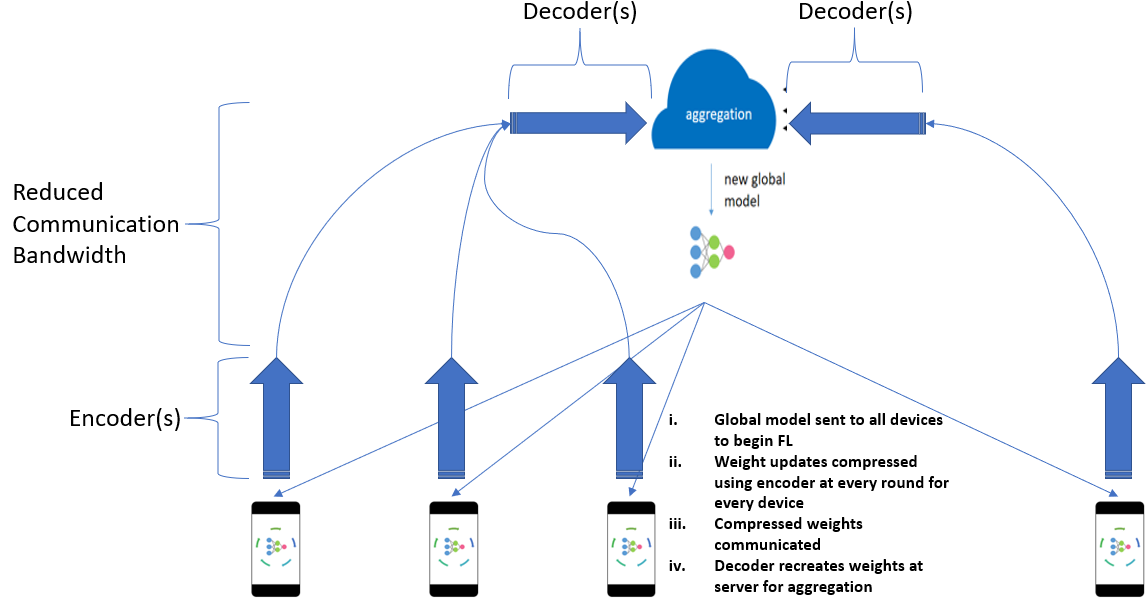}}
\caption{FL with AE compression}
\end{center}
\vskip -0.2in
\end{figure}

\section{Detailed Methodology}
\label{Methodology}
\subsection{Experimental setup}
\label{Experimental setup}
To test the AE-based weight compression on the weight updates of a training, analyses are performed on image classifier model weights by compressing the weights at the end of every epoch (analogous to communication round). The collaborator model (image classifier) is allowed to completely train, and the weights at regular intervals are stored in a dataset. This dataset is used to train the AE. The Fully Connected (FC) AE construct has dense layers and the input and output layer size is equal to the total number of parameters in the classifier model (or each weight update). The layers' size/shape of the AE is such that it follows the funnel structure represented in Fig. \ref{Fig AE}, decreasing as layers progress away from the encoder, reaching a minimum, and then subsequently increasing till the end of the decoder. This symmetric AE is then allowed to train on the classifier weights dataset generated earlier. 
\\
First, a fully connected AE approach is considered and the methodology is explained using the MNIST classifier as the collaborator model. The classifier model is comparatively simple with 15,910 parameters, and the convergence is reached in 10 epochs. To train an AE to replicate a model's weights, it is required to have a wide spectrum of data points in terms of weights at different instances of training, that can define the model uniquely to a good degree. 
\\ 
To simulate a more complex scenario, the next stage of analysis has a more complex classifier model, to generate a better and more widespread dataset in terms of weights. The CIFAR-10 classifier model has 550,570 parameters, and the convergence is reached in around 100 epochs. The weights being stored in the dataset, cannot be directly passed as inputs to the AE, given that it's of an unflattened structure. Since the flattening process of this dataset which has each element of size 550,570 is computationally challenging, the number of epochs is restricted to 40. This is done to put a cap on the dataset size, bearing in mind that a significant accuracy has been reached by this stage.  

\subsection{Dynamic AE architecture}
 The size of the AE is dependent on the size of the model whose weight updates are to be compressed. The first layer of the AE, by it having the weight updates from the collaborator model as inputs, has an input shape that matches the parameter size list (flattened single dimensional copy of the weights of the collaborator model). Unlike traditional compression algorithms, the compression methodology is dependent on the compressing item which is the collaborator model here. The complexity of the AE is also a feature that can be set according to the user needs, and infrastructure dependencies. Other than the first and last layer of the AE which has a size equal to that of the number of parameters in the collaborator model, the layer architecture can be varied to control the AE model complexity. By increasing the complexity, the AE mimics the learning of the collaborator model better. However, to reduce computational overheads, if accuracy can be compromised, then the complexity of the model can be reduced to meet the exact design requirements in terms of computation power usage vs accuracy trade-off for the given FL scenario. 
 \\
The compression ratio in AE-based compression is not pre-defined. The general observation is that as the compression ratio increases, more information loss exists during the recreation phase through the decoder end at the aggregator, and thus causes a dip in accuracy. This is because AE-based compression is inherently a lossy-compression technique that tries to encode the inputs into a smaller feature space by extracting certain relations between the parameters of the weight updates. Lossy compression is acceptable in the FL setup, because there is finally an aggregation of the weights, thus allowing for some room for difference in predictions and actual weights. Thus, this approach can be tailored to suit the needs of the particular FL scenario. 
\\
In many cases, the server requires an approximation of the weight updates from the collaborator, to propose the next set of weights for each collaborator. This is due to the aggregation process that ensures all weights are averaged together using different techniques. In such scenarios, a less complex AE could be used to save computational overheads of training the network on collaborator devices. The compression ratio could also be higher, if the accuracy obtained post lossy compression (and information loss), is sufficient for the aggregator function to suggest an appropriate update from aggregator to collaborator, for the next communication round. Alternatively, if higher accuracy is required for a particular FL set-up, a more complex AE is used to learn the relations between parameters of the weight updates better. The compression ratio may also be reduced to ensure lesser information is lost through the lossy compression done using AE. In addition, AE-based compression is an orthogonal approach and can be applied in combination with certain other traditional compression methods mentioned under the past work section. 

\subsection{Fully Connected AE limitation and proposing Convolutional AE as an alternative}
\label{Limitations}
The drawback of a fully connected AE-based compression is that the AE network is a  large model, with the number of parameters being a few hundred times bigger than the size of the original collaborator model in many cases. This comes with a computational overhead while training this network, and also a communication overhead while passing the decoder (half the size of the AE) at the end of the pre-pass round in FL. To achieve useful results of compression, the communication savings achieved by the AE compression ratio must be higher than the cost of communicating this decoder model at the end of the pre-pass round. This is possible if the number of collaborators and the number of communication rounds in the FL setup increase, as it would account for more communication overheads savings through AE compression. (Discussed in detail in subsequent sections.)
\\
An alternative approach is using a convolutional AE that will be significantly smaller in size when compared to an FC AE. The underlying logic behind the usage of convolutional AE is that weights across various layers are analogous to the pixel values in images. The dependency between parameters that the AE learns may be more evident between weights of layers that are closer to each other as compared to the dependency between weights of layers that are much further apart in the neural network. In this sense, a convolutional AE fits the context of compression better, as a convolutional network can learn the local (neighborhood) dependence better. A preliminary setup has been suggested for the same [Appendix]. The authors include this here to probe further research on convolutional AE with deeper networks or ones that implement residual networks, etc.

\section{Results and Analyses}
\label{Results}
\subsection{Fully connected AE}
\label{Valdation algorithm}

To analyze the performance of AE compression, beyond just its training metrics, a validation model is built as follows: once the AE has been trained, the weights logged at the end of every epoch from the original collaborator training are used to perform the validation. The trained AE is used to compress and predict (recreate) these weights. These AE predicted weights are then set as weights to another model following the same architecture as the collaborator model, and the loss and accuracy are computed by allowing the network to train for an epoch by ensuring the weights are fixed (trainable is set to false). The idea of this validation model is to show that the loss and accuracy values generated by setting the AE predicted weights to the collaborator model, are similar to the loss and accuracy values generated during the original training of the collaborator model. This would show that the AE has successfully learned the encoding of the collaborator model weights, performs compression, and recreates a set of weights that is similar to the original weights- which is seen through the fact that the loss and accuracy plots generated are similar.

\subsubsection{AE compression on MNIST Classifier}

\begin{figure}[ht]
\begin{center}
\centerline{\includegraphics[width=\columnwidth, height=4.5cm] {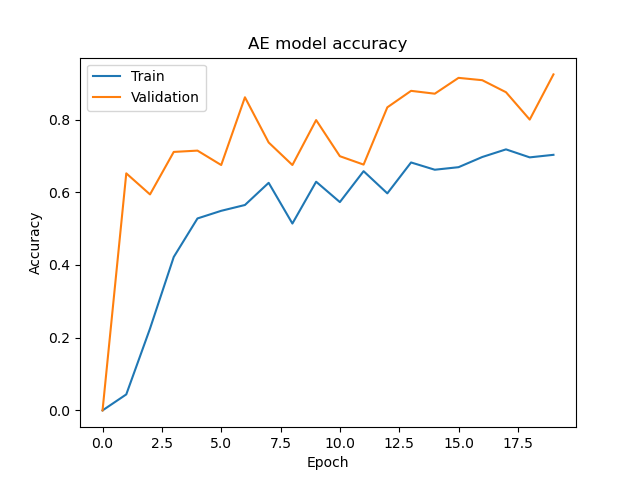}}
\caption{AE accuracy plot for MNIST classifier model compression}
\label{AE_mnist_acc}
\end{center}
\end{figure}

\begin{figure}[ht]
\begin{center}
\centerline{\includegraphics[width=\columnwidth, height=4.5cm]
{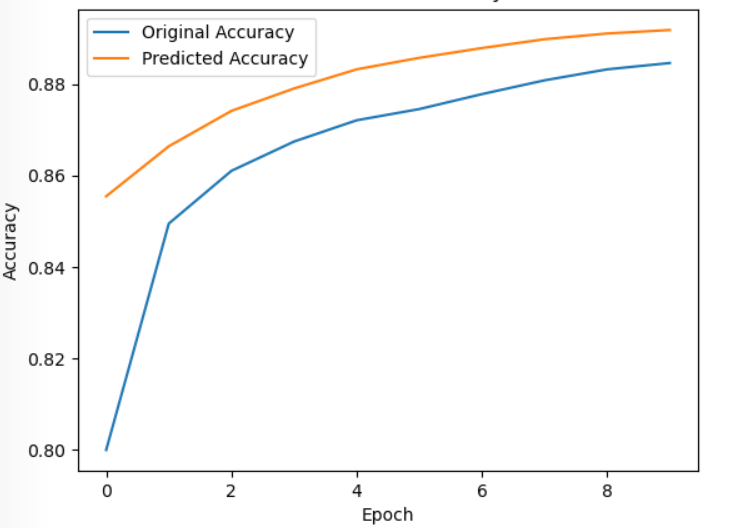}}
\caption{Accuracy variation when - MNIST model is set Original vs AE Predicted weights}
\label{mnist_validation_acc}
\end{center}
\end{figure}

Fig. \ref{AE_mnist_acc} shows the accuracy plot of the FC AE during training and as the loss converges. The FC AE can train accurately, as is seen with a maximum value of 0.78 in the accuracy metric (validation accuracy= 0.94). These results show that an AE (with 1,034,182 parameters) can compress and recreate the weight updates for a relatively simpler model such as an MNIST classifier. The results of the validation model (as proposed in the previous section) are shown next. As seen in Fig. \ref{mnist_validation_acc}, it is evident that the AE model can mimic the accuracy (and loss) plot of the original classifier training through the predicted (post-compression) weights. By the AE compression, the size of each update is reduced to a 32 feature encoding, and thus achieves about 500x compression. The reduced feature space size is a variable and can be modified to achieve varying levels of compression and accuracy.

\subsubsection{AE compression on CIFAR Classifier}
\begin{figure}[ht]
\begin{center}
\centerline{\includegraphics[width=\columnwidth, height=5cm]
{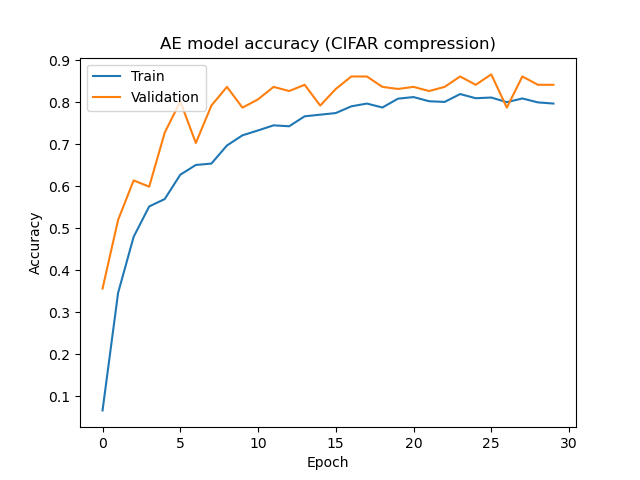}}
\caption{AE accuracy for CIFAR classifier model compression}
\label{AE_cifar_acc}
\end{center}
\end{figure}

\begin{figure}[ht]
\begin{center}
\centerline{\includegraphics[width=\columnwidth, height=5cm]
{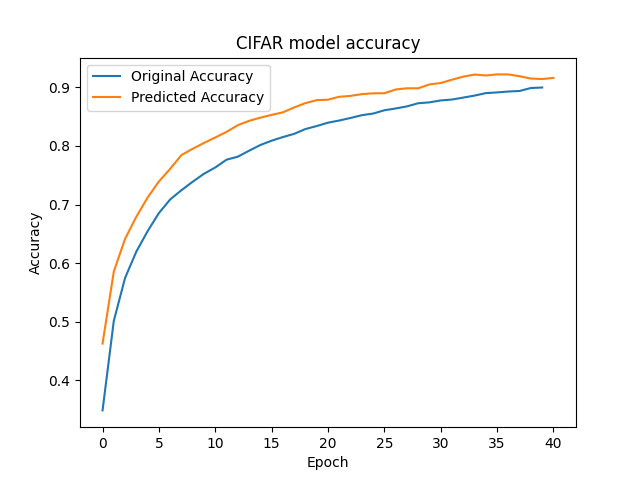}}
\caption{Accuracy variation when - CIFAR model is set Original vs AE Predicted weights}
\label{cifar_validation_acc}
\end{center}
\end{figure}

The next analysis is on a more complex collaborator model. The CIFAR-10 classifier model is more complex and serves as the proof of concept for the AE-based compression method.
This model also has significantly more parameters (352,915,690) and more layers. The weights in the dataset have a higher variance and suit better for AE training as is seen in Fig. \ref{AE_cifar_acc}. The AE learns the encodings and mimics the inputs successfully as is seen through the accuracy plot. The highest accuracy reached is around 0.79 (validation accuracy= 0.83) and the loss converges after around 25 epochs. The results of the validation model also show that the loss and accuracy metrics are similar- when a CIFAR classified model has been set with the original weights, and with the AE predicted weights. This shows that while the AE compression is lossy, it doesn't affect/change the weights recreation at the server end to an extent where the corresponding accuracy (and loss) changes drastically. This shows that the AE is replicating the learning properties of the original collaborator training Fig.\ref{cifar_validation_acc}. The predicted accuracy is higher (even for MNIST) due to finetuning during prepass AE construction. Increasing the threshold between collected weights during prepass can reduce this.

\subsection{Testing the Fully-Connected AE compression in an FL setup}
\begin{figure}[ht]
\begin{center}
\centerline{\includegraphics[width=\columnwidth, height=5cm]
{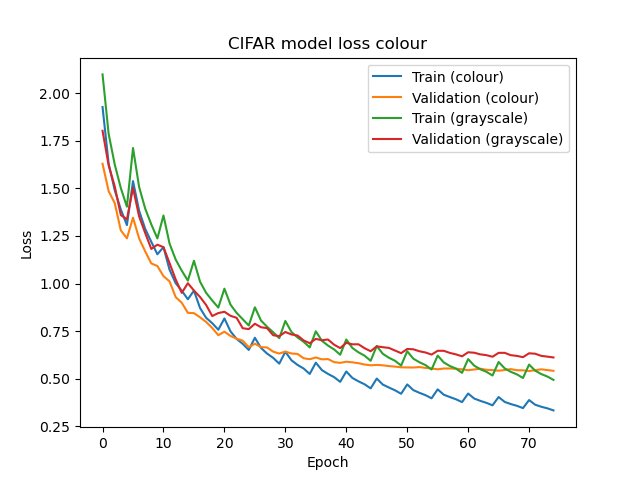}}
\caption{Loss for 2 collaborators (colour and grayscale classifier) in FL}
\label{FL_loss}
\end{center}
\end{figure}

\begin{figure}[ht]
\begin{center}
\centerline{\includegraphics[width=\columnwidth, height=5cm]
{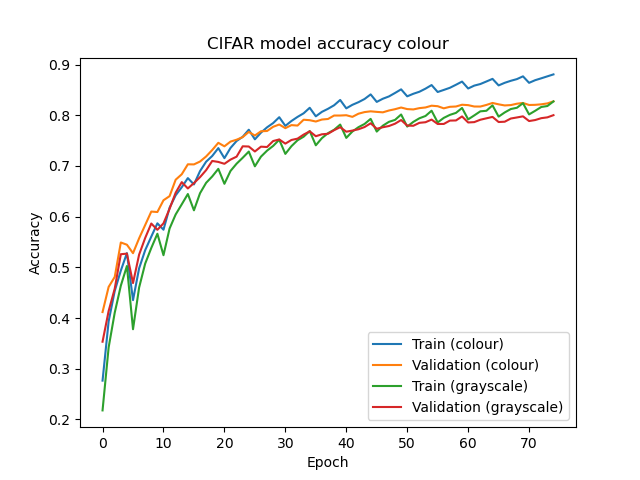}}
\caption{Accuracy for 2 collaborators (colour and grayscale classifier) in FL}
\label{FL_acc}
\end{center}
\end{figure}

 Overfitting decreases however with an increase in AE model complexity, or with a more complex model showing a greater gradient of weights through training. Overfitting as such does not pose a threat as it does in traditional ML, as the important task here is to finally compress and recreate the weights. However, in some cases, the weight updates from the collaborator that have been based on the previous round aggregations may vary significantly causing the AE to perform poorly. Here the dynamic AE model is to be modified accordingly, to increase complexity or reduce the compression ratio, to ensure the AE model is recreating the weights accurately. 
To validate this, a two-collaborator FL setup with color imbalances has been built, and the AE compression has been implemented. One collaborator is a CIFAR-10 classifier model on color images, while the other is on grayscale images (Colour Imbalance in datasets for each collaborator). There are 40 communication rounds, and each round has a local training running for 5 epochs. At the end of every communication round, the converged weights from both the collaborators are passed through their respective AE (compressed - communicated - recreated). A simple averaging-based aggregation algorithm is used to propose the global update for the next communication round.
The graphs below (Fig. \ref{FL_loss} and  \ref{FL_acc}) validate this, as is seen in the sawtooth plot for accuracy (or loss). The dips occur at the start of every communication round due to aggregation. These results show that the AE compression works for the proposed FL setup, as both the collaborators train accurately, despite being compressed by a factor of 1720x at the end of every communication round. (AE properties similar to previous sections)

\subsection{Dynamic compression- trade-off and break-even point for Fully-Connected AE}
\label{Savings ratio}
AE-based compression is dynamic as it works managing variables such as computational capacity, compression ratio, accuracy requirements, and other such dependencies in the FL setup. The graphs below show the trade-off and the respective break-even points between the cost of communicating the decoder model(s) at the end of the pre-pass round and the communication savings achieved through AE compression. There are 2 conditions shown- a) When one decoder model suffices for the entire federated learning set-up, and b) where each collaborator has a decoder model. The savings achieved while assuming the single decoder condition are dependent on the number of communication rounds and the number of collaborators. Similarly, the savings achieved in the multiple decoder approach, is dependent only on the number of communication rounds, as the number of collaborators is a factor present in both the cost and the saving. Conditions (a) and (b) are extreme cases, and there may be a scenario where a particular FL setup may require few decoders (less than the number of collaborators). For each of these cases, the AE model has 352,915,690 parameters and achieves nearly 1720x compression.  The Savings Ratio, in general, is calculated according to this equation:

\begin{equation}
\small
SR= \frac{Original Size \times Comm Rounds  \times Collabs}{Compressed Size \times Comm Rounds  \times Collabs + Cost}
\end{equation}
where- Original Size is the actual model size,
Compressed Size is the compressed model size,
Comm Rounds are the number of communication rounds,
Collabs are the number of collaborators, and
Cost is the Decoder overheads
\begin{equation}
\small
Cost= Decoder Size\times No. of Decoders
\end{equation}
\begin{equation}
\small
Cost= \frac{Autoencoder Size}{2}\times No. of Decoders
\end{equation}

\begin{figure}[ht]
\begin{center}
\centerline{\includegraphics[width=\columnwidth, height=5cm]
{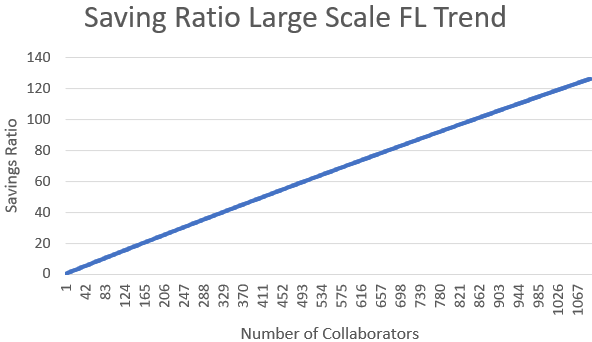}}
\caption{Case (a) Savings Ratio- 120x Savings Ratio beyond 1000 collaborators, breakeven point at 40 collaborators}
\end{center}
\end{figure}

\begin{figure}[ht]
\begin{center}
\centerline{\includegraphics[width=\columnwidth, height=5cm]
{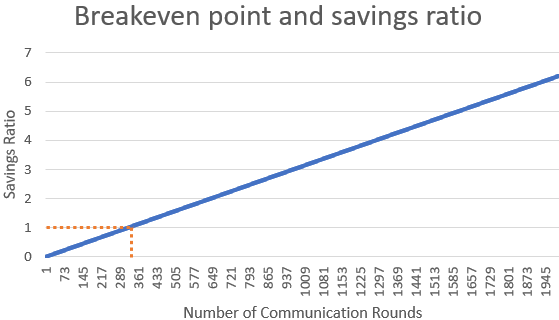}}
\caption{Case(b) Breakeven point when No. of Comm rounds= 320, SR improves for large scale FL}
\end{center}
\end{figure}

\section*{Conclusion}
The dynamic and orthogonal nature that can set the exact compression ratio makes this approach one that performs the maximum compression compared to any other compression model. The caveat being that there could be a situation where the loss of information becomes high. This compression allows for the preservation of accuracy, by increasing the complexity of the AE, at the expense of computational costs. Thus, a trade-off-based analysis proves to be useful while employing an AE-based compression. This paper considers generic collaborators models while performing the AE compression and shows that compression ratios achieved can go nearly as high as 2000x, while also maintaining required levels of accuracy. It investigates the efficacy of AE compression for generic models in the FL setup and shows that collaborator models can successfully train, while the updates are being AE compressed at every communication round. Finally, trade-off analyses show that the advantages are greater as the number of collaborators and communication rounds increase.

\bibliographystyle{named}
\bibliography{ijcai21}

\end{document}